\documentclass[letterpaper, 10 pt, conference]{IEEEtran}

\let\labelindent\relax

\usepackage[letterpaper, top=22.1mm, bottom=22.1mm, left=22.1mm, right=22.1mm]{geometry}

\usepackage{graphicx}
\usepackage{mathtools, nccmath}
\usepackage{multirow}
\usepackage{algorithm,algpseudocode}
\usepackage{booktabs}
\usepackage{gensymb}
\usepackage{comment}
\usepackage{amsfonts}
\usepackage{subfig}

\usepackage[numbers]{natbib}
\usepackage[dvipsnames]{xcolor}

\usepackage[font=footnotesize]{caption}
\usepackage{balance}

\usepackage{gensymb}

\newcommand{\cmmnt}[1]{}

\usepackage{titlesec}
\titlespacing{\section}{0pt}{*0.3}{*0.3
}
\titlespacing{\subsection}{0pt}{*0.2}{*0.2}
\let\labelindent\relax
\usepackage{enumitem}
\setlist[itemize]{noitemsep, ,nolistsep,topsep=0pt}
\setlist[enumerate]{noitemsep,nolistsep, topsep=0pt}

\usepackage{pgfplots}
\usetikzlibrary{external, positioning}
\usepackage{pgfplots}
\pgfplotsset{compat=1.9}
\usepgfplotslibrary{statistics}
\usetikzlibrary{shapes}
\usetikzlibrary{er}

\IEEEoverridecommandlockouts                   

\title{\LARGE \bf Visualizing Robot Intent for Object Handovers with Augmented Reality}
\author{Rhys Newbury, Akansel Cosgun, Tysha Crowley-Davis, Wesley P. Chan, Tom Drummond, Elizabeth A. Croft \\ Monash University, Australia
}

\begin{document}

\maketitle
\thispagestyle{empty}
\pagestyle{empty}

\begin{abstract}

Humans are highly skilled in communicating their intent for when and where a handover would occur. However, even the state-of-the-art robotic implementations for handovers typically lack of such communication skills. This study investigates visualization of the robot's internal state and intent for Human-to-Robot Handovers using Augmented Reality. Specifically, we explore the use of visualized 3D models of the object and the robotic gripper to communicate the robot's estimation of where the object is and the pose in which the robot intends to grasp the object. We tested this design via a user study with 16 participants, in which each participant handed over a cube-shaped object to the robot 12 times. Results show communicating robot intent via augmented reality substantially improves the perceived experience of the users for handovers. Results also indicate that the effectiveness of augmented reality is even more pronounced for the perceived safety and fluency of the interaction when the robot makes errors in localizing the object.

\end{abstract}

\section{Introduction}
\label{sec:introduction}

Object handover is a ubiquitous interaction type between humans, and an important skill for robots that interact with people. Humans often communicate their intent for when and where a handover will occur using several modalities, including speech, gaze, or body gestures. This observation suggests that robots will also require such communication skills. Recent years have seen a proliferation of research in human-robot handovers~\cite{handovers_patrick, kwan2020gesture, ardon2020affordance, handovers_yang}. While communication cues such as speech and gaze positively impact human-robot handovers~\cite{moon2014meet}, a recent survey found that a majority of robotic systems that perform handovers did not use any communication cues at all~\cite{ortenzi2021object}. These findings suggest that there is a need for innovation in how to communicate intent for human-robot handovers.

Recent advancements in graphics and hardware technology have led to increased popularity of Augmented Reality (AR). AR presents new opportunities for robotics applications, and it is especially promising for Human-Robot Interaction~\cite{AR_Review}. Using AR, robots can communicate their intent to users through visual displays, allowing users to consider the robot's plans and act accordingly. This idea has found some success in previous studies where AR was used to communicate the future motion of robotic arms~\cite{intent_arm, intent_arm2}.

\begin{figure}
    \centering
    \includegraphics[trim= 0 200 400 0,clip,width=0.965\linewidth]{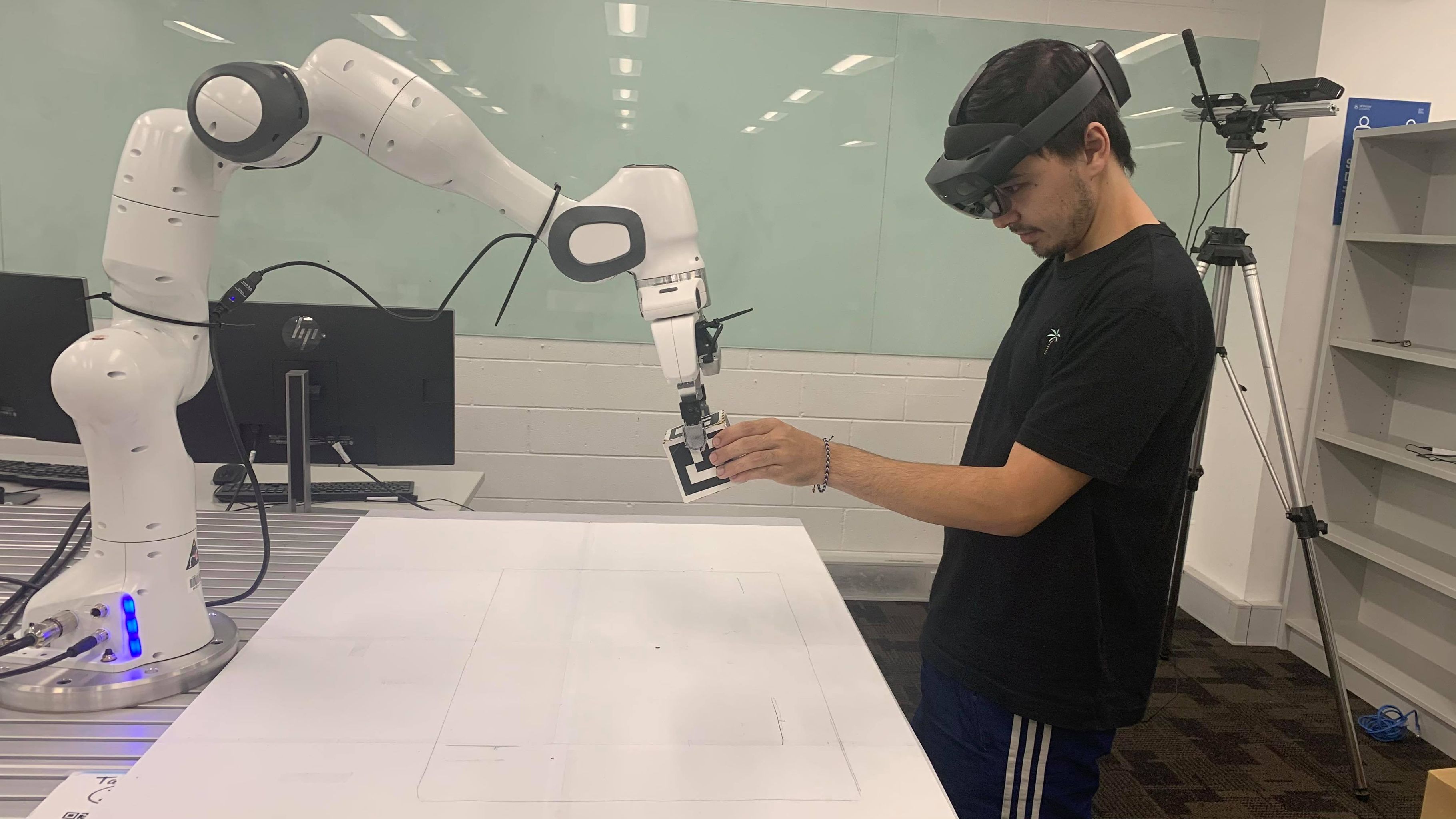}
    
    \includegraphics[width=0.48\linewidth]{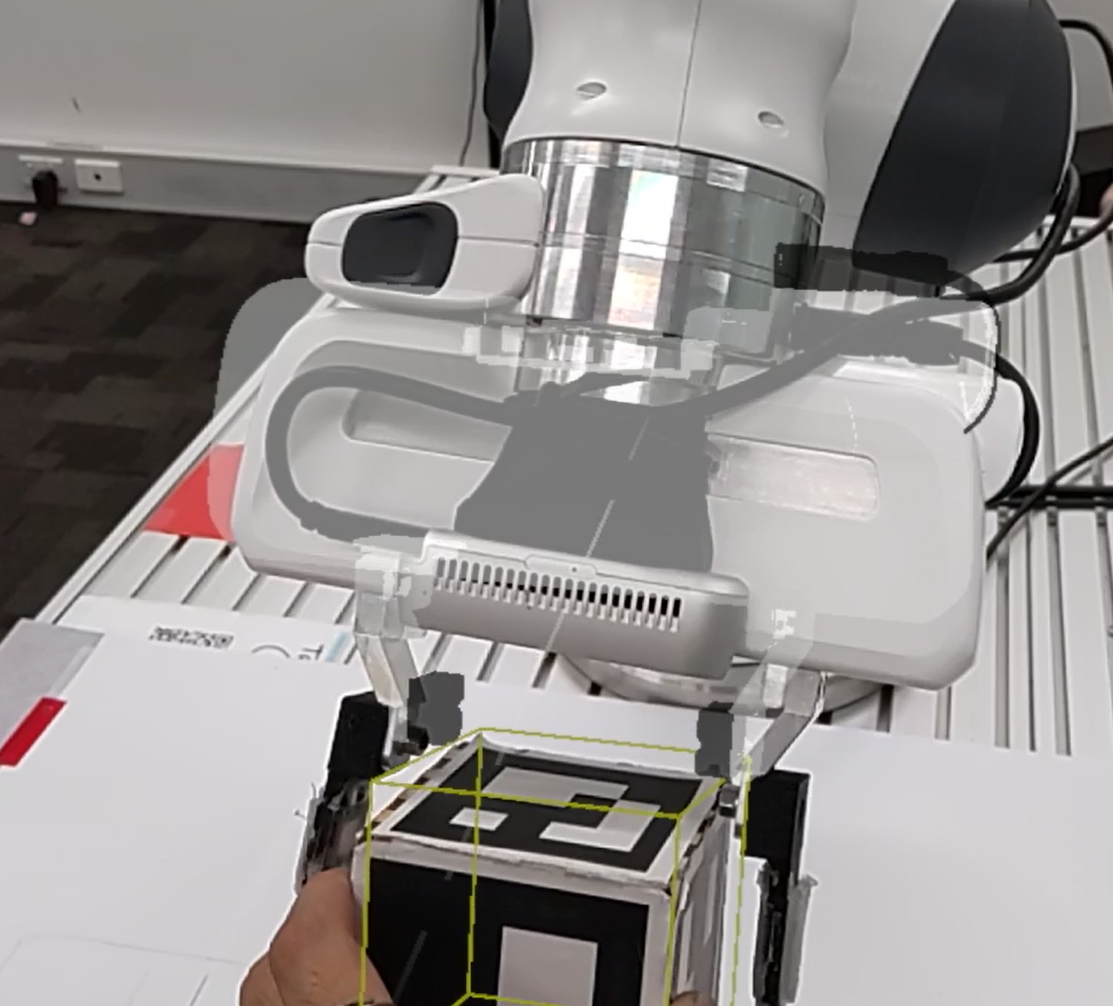}
    \includegraphics[trim=250 0 455 200, clip,width=0.48\linewidth]{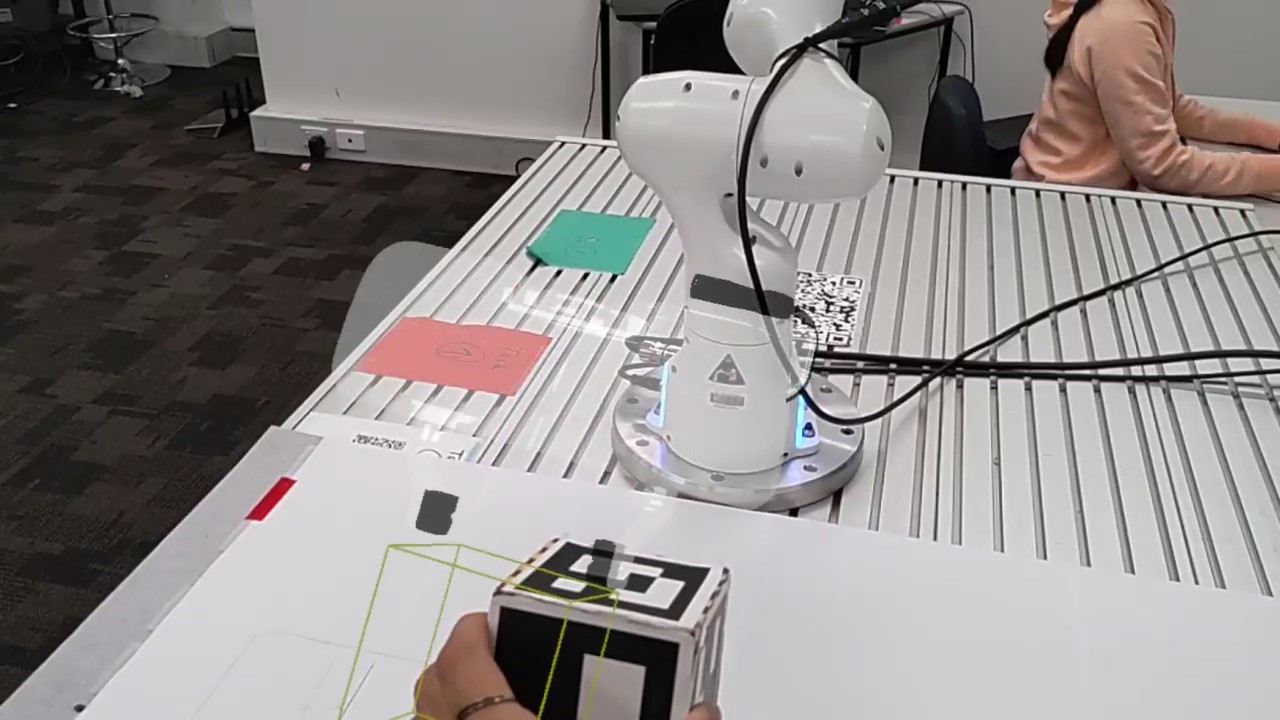}

    \caption{Top: The robot picking up an object from the user's hand. The user is wearing an Augmented Reality headset. Bottom Left: The detected pose of the object, and AR visualization of how the robot is planning to grasp the object. Bottom Right: A simulated error is introduced to the object pose estimation, to test how the system would perform when the robot makes errors.}
    \label{fig:intro}
    \vspace{-0.25cm}
\end{figure}

In this paper, we explore the use of AR for a specific application: to communicate the robot's internal state and intent in Human-to-Robot handovers. We propose visualizing two 3D models through AR: 1) The detected object pose, visualized as a wireframe of the object model, and 2) The planned grasp pose, visualized as a virtual, low-opacity 3D model of the robotic gripper. A single object, which is tracked with the help of artificial markers, is used for the experiments. Figure~\ref{fig:intro} shows the experimental setup and example views from the AR Headset. We conduct a user study to compare the perceived safety, trust and fluency of the handovers using our proposed AR interface. Half of the conditions for each user involved adding a random but controlled \textbf{simulated error} to the robot's estimation of the object pose to simulate markerless vision-based pose tracking in which accurate pose estimation is challenging. \\
The contributions of this paper are two-fold:
\begin{itemize}[topsep=0pt]
    \item To the authors' knowledge, this is the first reported use of Augmented Reality for human-robot handovers, to convey the robot's belief of where the handover would occur, and its plan to grasp the object.
    \item User studies that confirm the effectiveness of the proposed AR communication approach.
\end{itemize}

\section{Related Work}
\label{sec:related_work}


\subsection{Human-to-Robot Handovers}
As the need for collaborative manipulation increases, the importance of researching object handovers between robots and humans continues to grow. A recent survey paper by \citet{ortenzi2021object} reviews the progress made in this field. They summarize different capabilities that enable handovers in the robotic system, including communication, grasp planning, perception, and error handling. Among the papers surveyed, most focus on Robot-to-Human handovers, while only a small fraction of the papers focus on Human-to-Robot handovers. The authors observe that communication during handovers is often overlooked, with very few papers primarily focusing on the communication of intent. This highlights a gap in the research regarding how to communicate the robot's intent to the user effectively.


Literature in the field has predominantly focused on specific aspects of handovers, such as the trajectories of the arm either learned from human-data~\cite{human_traj1}, constructed using dynamic primitives~\cite{dyn_prim} or predicted from the transfer point of the object~\cite{Nemlekar}. Two recent works focus on creating a human-to-robot handover system that can generalize to a series of objects~\cite{handovers_patrick, handovers_yang}. These approaches use a deep learning-based grasp planner and skin segmentation to find a safe grasp pose for the robot. 

\subsection{Communication of Robot Intent}
\citet{norman1988psychology} proposed the idea of the \textit{Gulf of Evaluation} as the relative ability of the user to directly interpret the state of the system. In this context, during human-robot interaction, a lack of communication of goals and intent from the robot can lead to a large gulf. Effective communication from the robot can enhance the user's perception of the reliability of the system and make the human feel more comfortable around the robot~\cite{onmymind}.

One option to achieve a more natural communication between the human and the robot is via a head-mounted display (HMD) to visualize the robot's future motion from the point of view of the human~\cite{ruff, scass}. This idea has been utilized for both the motion of wheeled robots~\cite{intent_motion} and robotic arms~\cite{intent_arm, intent_arm2}. \citet{intent_motion} utilized AR to communicate the motion intent of a robot. They used different visualization markers to show the robot's intended path and directions, allowing the humans to know the robot's motions in advance. They found that AR could improve task efficiency over a baseline in which users only saw physically embodied orientation cues. \citet{intent_arm} utilizes AR to visualize the path of the robotic arm to allow better collaboration between human and robot. They show an increase in accuracy and decrease in the time taken to label a trajectory as either collision or collision-free with blocks on a table. \citet{intent_arm2} shows that using AR to visualize the robot's future motion can allow humans and robots to work in small shared spaces with a decreased likelihood of shutdowns. 

\subsection{Communication of Robot Intent in Handovers}

Research has previously explored the use of nonverbal cues in human-robot handovers. The use of gaze results in faster object reaching and more natural perception of the interaction by human receivers~\cite{moon2014meet}. \citet{admoni2014deliberate} further showed that modulating the speed of the handovers, such that the object released is delayed until the human gaze is drawn back to the robot, can increase the conscious perception of the robot's communication. Furthermore, integrating the use of both head orientation and eye gaze into the decision-making of the robot significantly increases the success rate of robot-to-human handovers~\cite{grigore2013joint}. Other communication modalities explored include body gestures, such as an extended arm which presents the object to the receiver such that the free part of the object is directed towards the receiver can convey intent to initiate a handover~\cite{cakmak2011using}. \citet{pan2018exploration} showed that the initial pose of the robot could inform the giver about the geometry of the handover and improve the fluency of the handover.  However, to our knowledge, AR has not been utilized for communicating robot intent in human-robot handovers.


\section{Approach}
\label{sec:system}

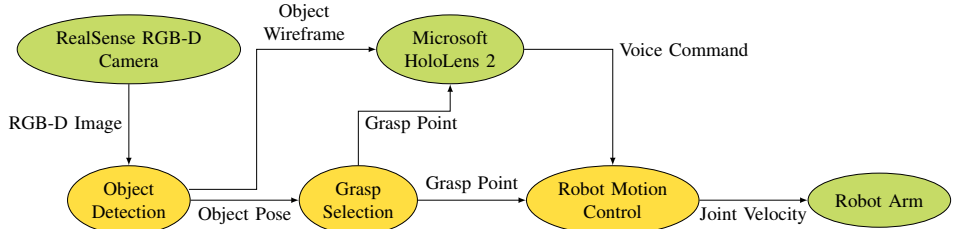
\begin{figure*}[h!]
    \centering
    \resizebox{.75\textwidth}{!}{
    \begin{tikzpicture}
 
 

\node[draw, 
      thin,
      draw=black,
      ellipse,
      minimum width=2cm,
      minimum height=1cm,
      align=center,
      fill=SpringGreen] (camera) at (0,0) {RealSense RGB-D \\ Camera};

\node [draw,
    fill=SpringGreen,
    minimum width=2cm,
    thin,
    draw=black,
    ellipse,
    minimum width=2cm,
    minimum height=1cm,
    align=center,
    minimum height=1.2cm,
    right=2.5cm of camera
]  (hololens) {Microsoft\\HoloLens 2};
 
 
\node [draw,
    fill=Goldenrod, 
    thin,
    draw=black,
    ellipse,
    minimum width=2cm,
    minimum height=1cm,
    align=center,
    below= 1.5cm of camera
]  (od) {Object\\Detection};

\node [draw,
    fill=Goldenrod, 
    thin,
    draw=black,
    ellipse,
    minimum width=2cm,
    minimum height=1cm,
    align=center,
    right= 2cm of od
]  (gs) {Grasp\\Selection};

\node [draw,
    fill=Goldenrod, 
    thin,
    draw=black,
    ellipse,
    minimum width=2cm,
    minimum height=1cm,
    align=center,
    right= 2cm of gs
]  (mc) {Robot Motion \\ Control};

\node [draw,
    fill=SpringGreen, 
    thin,
    draw=black,
    ellipse,
    minimum width=2cm,
    minimum height=1cm,
    align=center,
    right= 2cm of mc
]  (ra) {Robot Arm};

\draw[-latex] (camera.south) -- (od.north)
    node[midway,left]{RGB-D Image};
    
\draw[-latex] (od.east) -- (gs.west)
    node[midway,below]{Object Pose};
 

\draw [-latex] (od.east)++(-0.03,0.2) --  node[anchor=east, above=2pt ]{} ++(1.25,0) |- (hololens.west)node[above,at start, align=center, anchor=south west, yshift=75pt, xshift=0pt]{Object\\Wireframe};
 
\draw [-latex] (gs.north) --  node[anchor=east, above=2pt ]{} ++(0,1) -| (hololens.south)node[above,at start, align=center, anchor=south west, yshift=-15pt]{Grasp Point};

    
\draw[-latex] (gs.east) -- (mc.west)
    node[midway,above]{Grasp Point};
    
 \draw[-latex] (hololens.east) -| (mc.north)
    node[midway,right]{Voice Command};
 
 \draw[-latex] (mc.east) -- (ra.west)
    node[midway,below]{Joint Velocity};

    
 
 
 
 
 

\end{tikzpicture}
 
    }
    \caption{The system diagram. Hardware components are shown in green. The yellow software blocks were developed in this system..}
    \label{fig:system_diagram}
\end{figure*}

Our system consists of 3 modules, as shown in Fig. \ref{fig:system_diagram}. For simplicity, we use a generic cube object for handovers. We adopt a fixed world frame, with the origin located on the robotic arm's base, where gravity acts in the $-z$ direction. We estimate the pose of the cube by using the artificial markers located on each face of the cube. To pick up the object from the user's hand, a feasible grasp pose is selected among several predefined grasp poses with respect to the object frame. A voice command initiates the handover detected through an AR Headset. A simple servo controller is used to drive the robot end-effector on a linear path that connects the starting end-effector pose to the selected grasp pose. The linear path of the robotic end-effector aims to make the motion more reliable and predictable for the human. The robot's estimation of the object pose and the selected grasp pose is visualized using the AR Headset to give users a sense of the robot's intention.



\subsection{Hardware Setup}

We use a table-mounted Franka Emika Panda robotic arm. The robot has 7 degrees of freedom and a two-finger parallel gripper. To increase the robustness of grasps, we use custom-made gripper fingers made of silicon rubber. Our system uses a single Intel RealSense D435 RGB-D camera that is mounted to the end-effector. 

We use a Microsoft HoloLens 2 as the designated AR Headset, which provides accurate positional tracking of the user's head, allowing for accurate visualizations in 3D space. The coordinate frame transformation between the robot and the HoloLens is initially established using an artificial marker placed in a predefined location. After the initial calibration, the HoloLens continually updates its pose with respect to the robot base frame. Communication of data to the HoloLens is achieved using WiFi over a local area network.

\subsection{Object Detection}
\label{sec:object_detection}

An 8cm edge cube object was used for our experiments. We placed unique arUco markers on each side of the cube, with each arUco marker oriented such that the z-axis of the marker is pointing outwards. The pose of the cube can be calculated by detecting a single arUco marker and projecting a fixed distance in the $-z$ direction of the detected marker. When multiple markers are detected, the best fit for the pose is used. A low-pass filter is applied to the resulting pose to smooth any sudden changes in object pose.   

\subsection{Grasp Selection}
\label{sec:grasp_selection}

\begin{figure}
    \centering
    \vspace{0.2cm}
    \includegraphics[trim=1750 650 1800 1150, clip, width=0.105\linewidth]{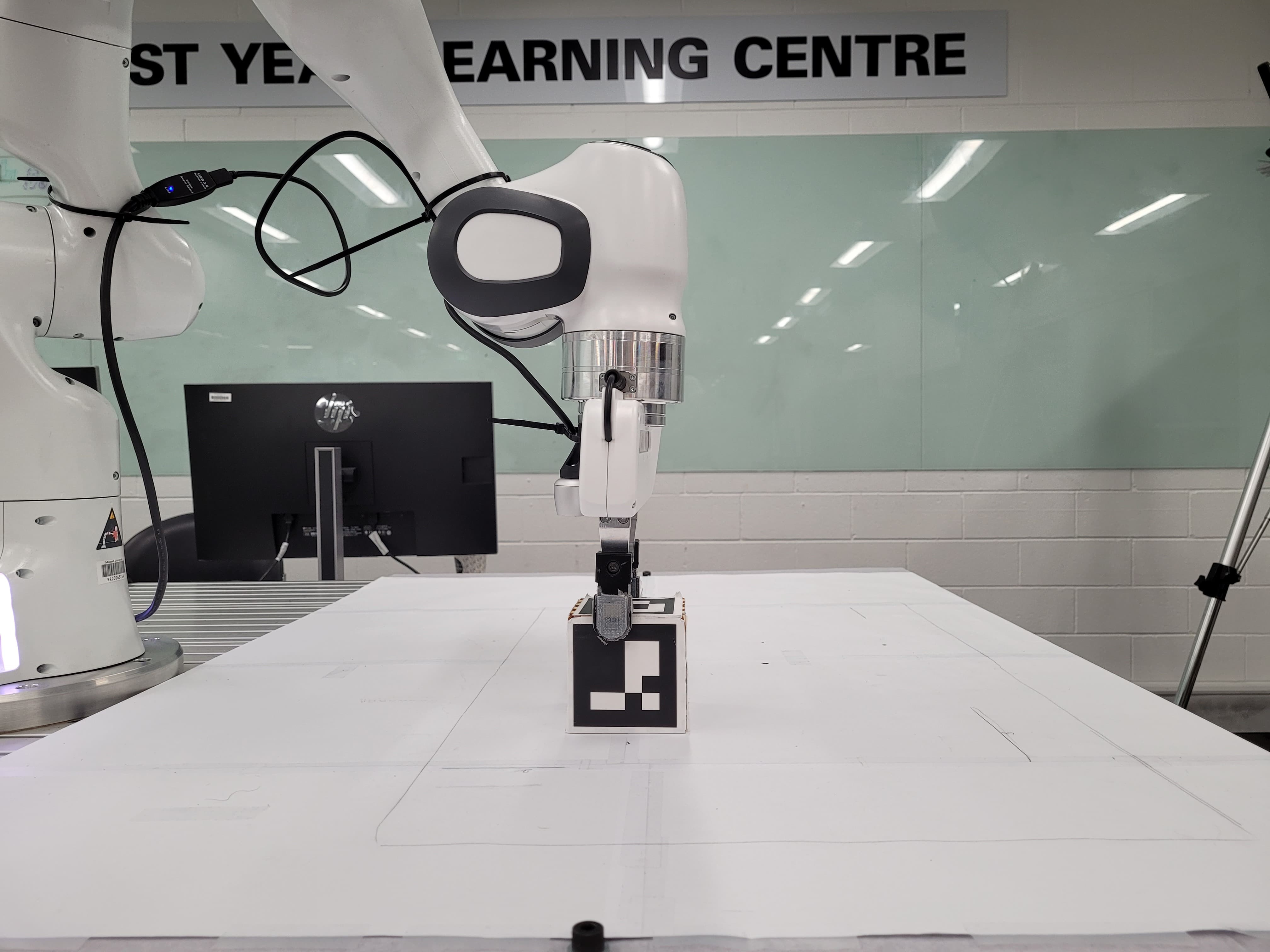}
    \hspace{-0.22cm}
    \includegraphics[trim=1370 650 1700 1150, clip, width=0.21\linewidth]{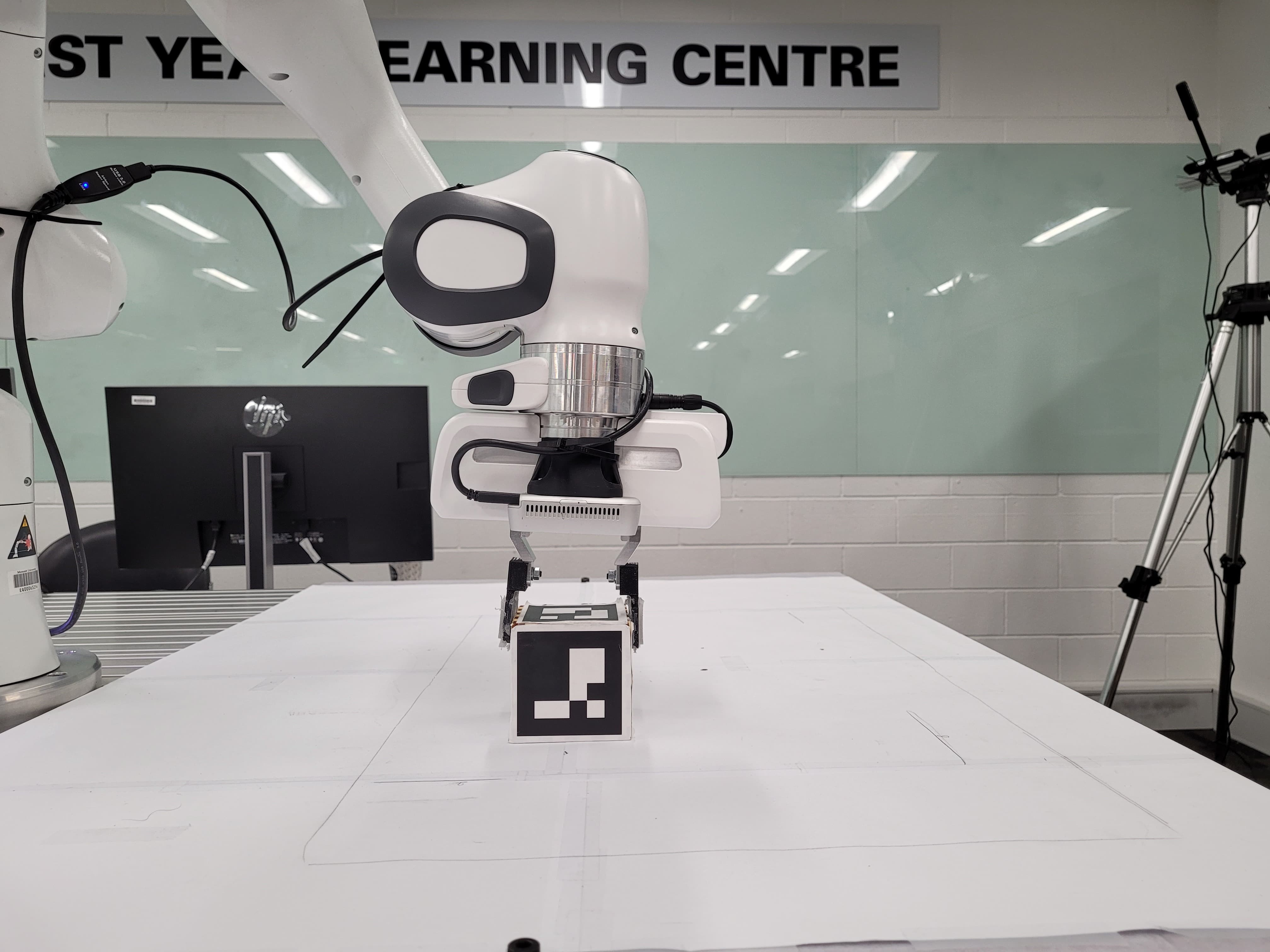}
    \hspace{-0.22cm}
    \includegraphics[trim=1750 650 1800 1150, clip, width=0.105\linewidth]{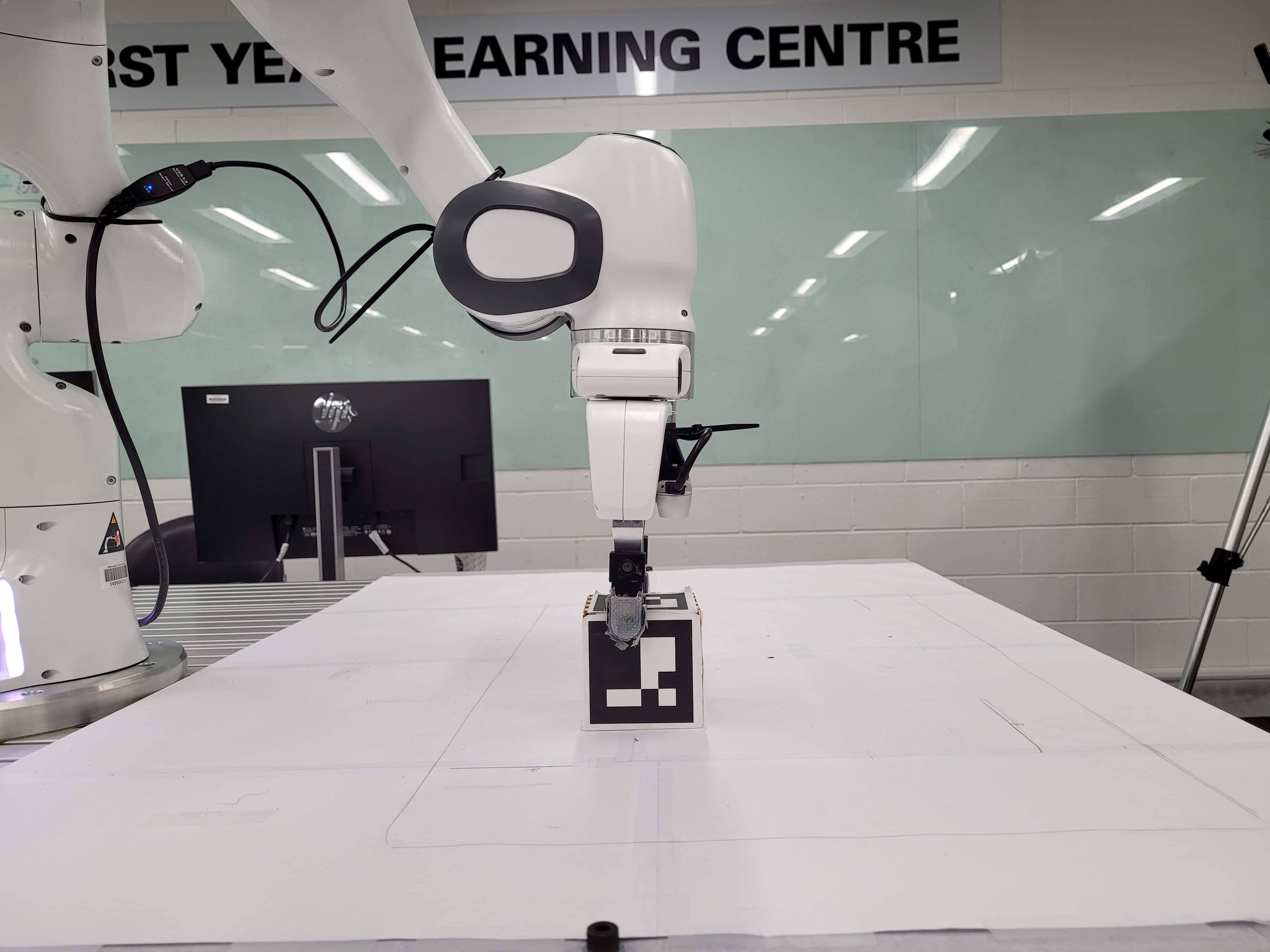}
    \hspace{-0.22cm}
    \includegraphics[trim=1360 650 1700 1140, clip, width=0.21\linewidth]{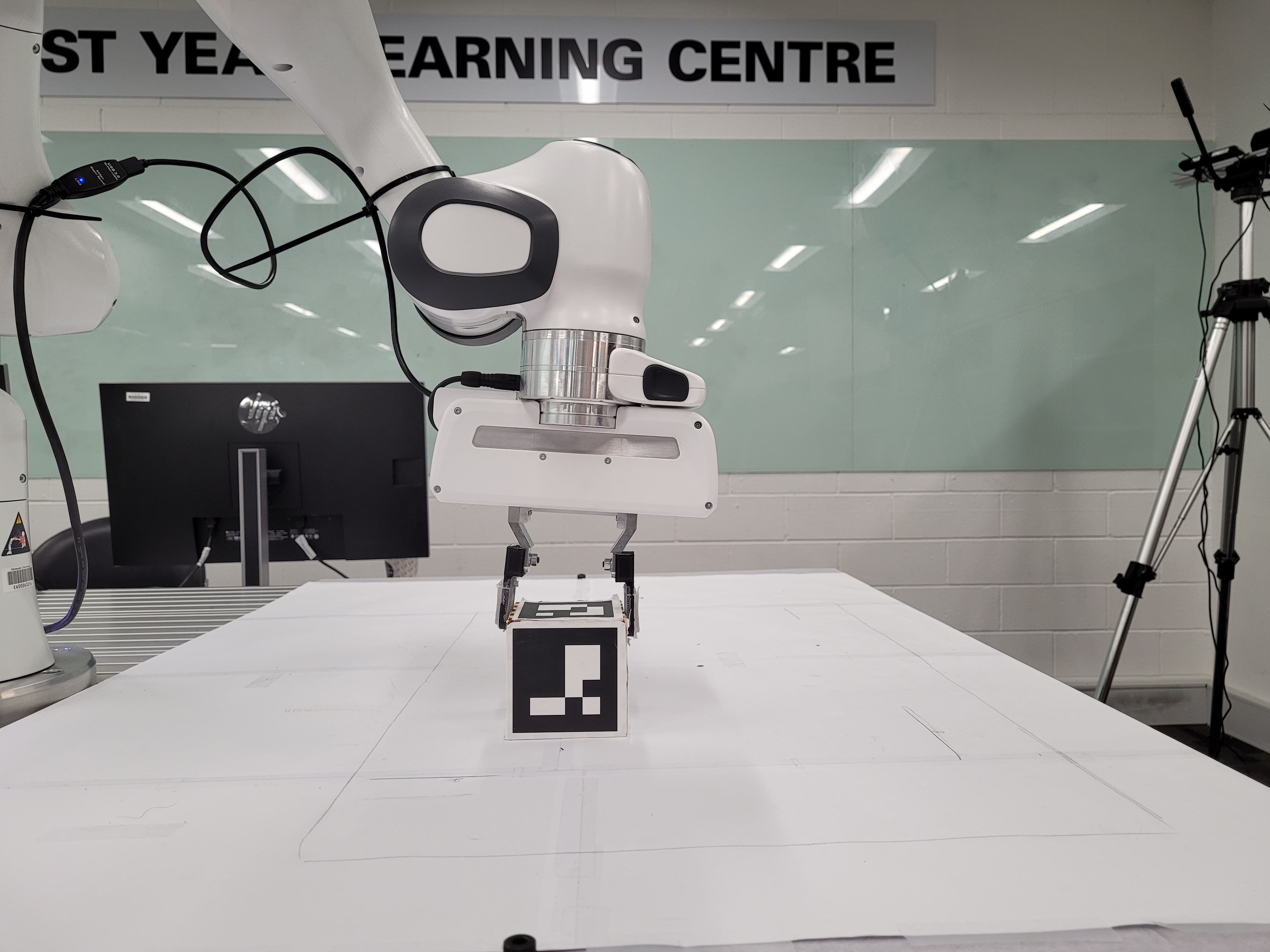}
    \hspace{-0.22cm}
    \includegraphics[trim=1770 650 1460 1250, clip, width=0.19\linewidth]{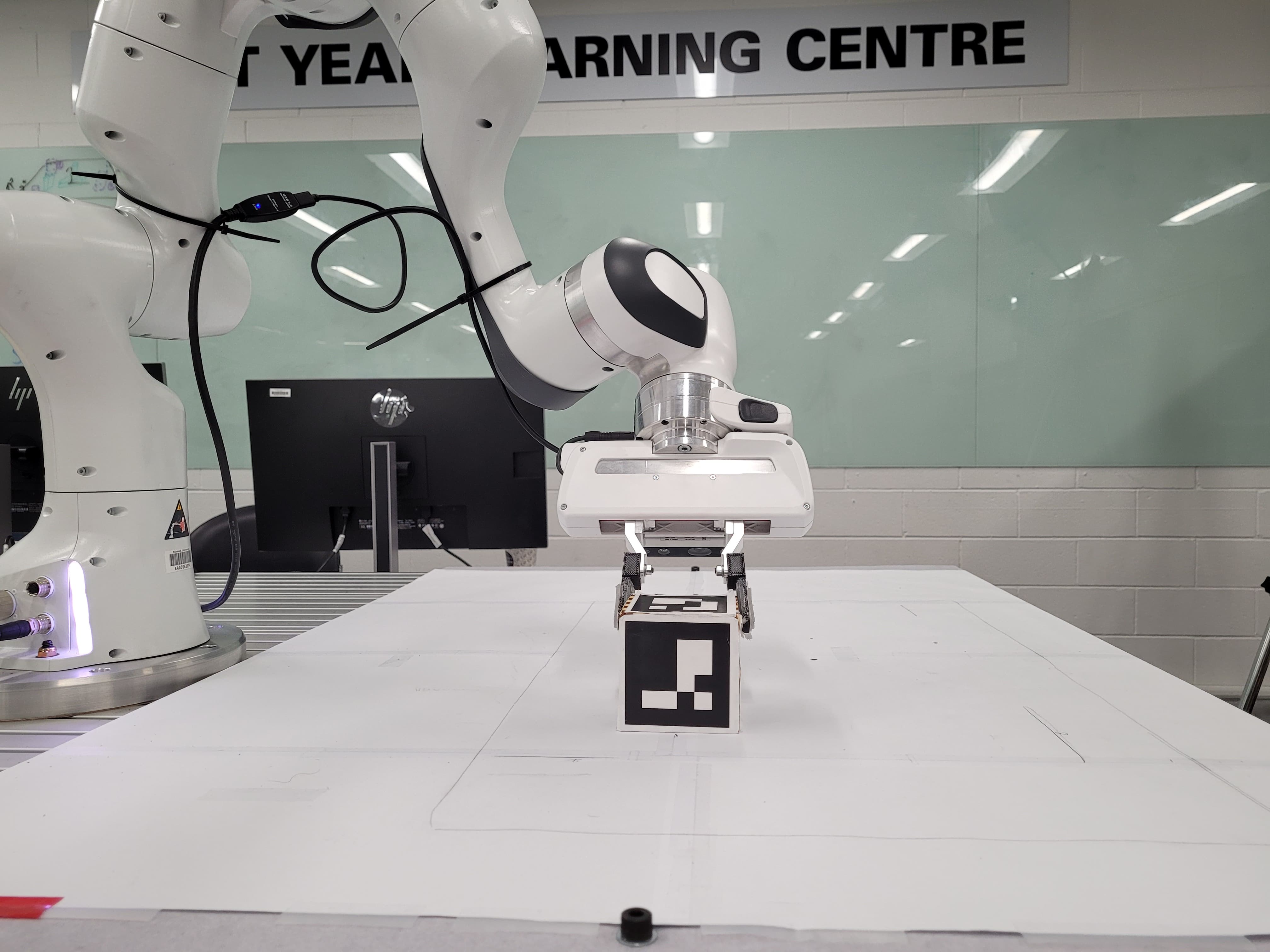}
    \hspace{-0.22cm} 
    \includegraphics[trim=900 340 715 590, clip, width=0.19\linewidth]{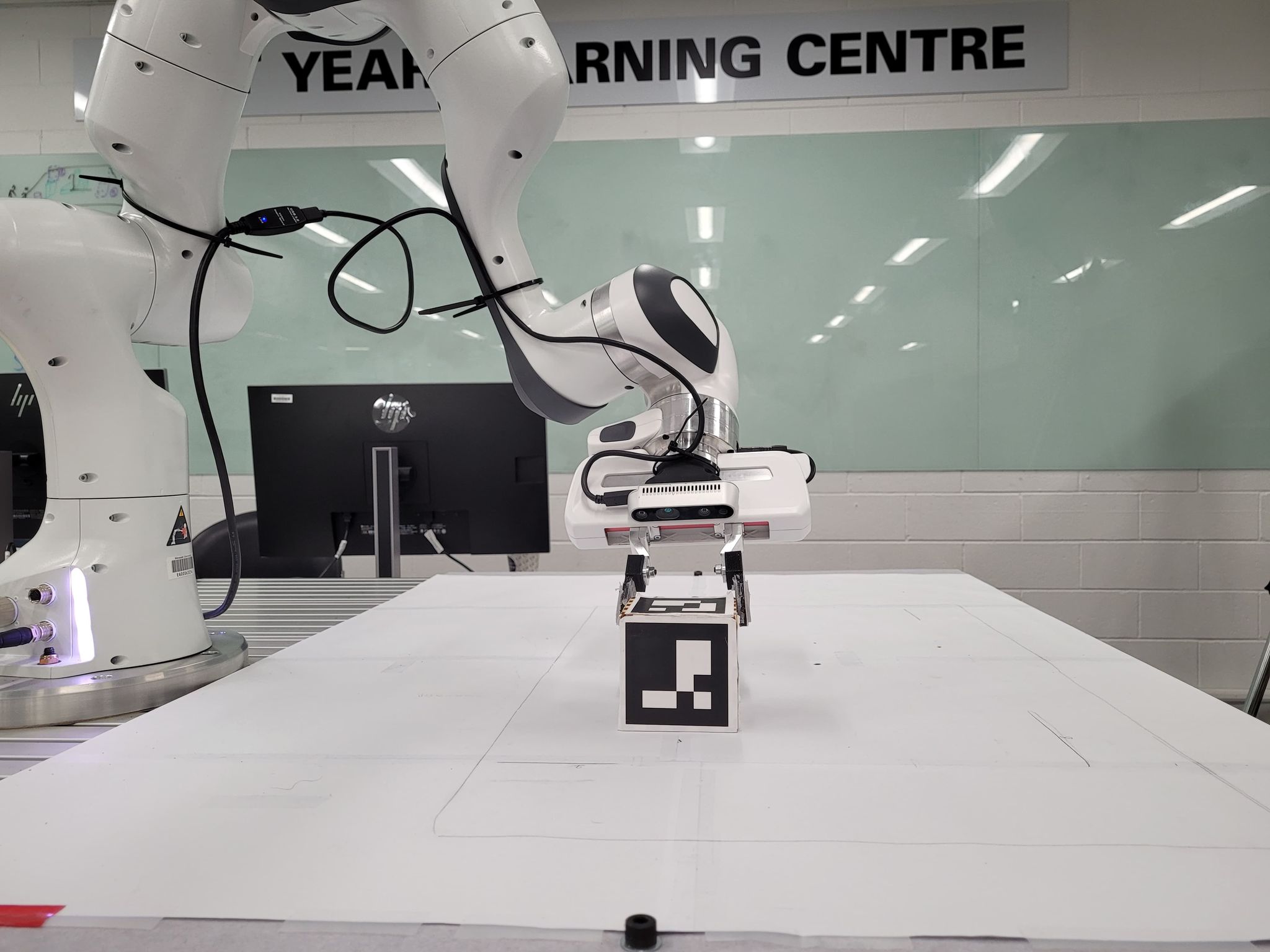}
    \hspace{-0.22cm}
    \caption{Predefined grasp poses for one face and one edge of the cube used in the experiments. To grasp a face of the cube, we consider 4 possible rotations of the end-effector that are all 90\degree apart. To grasp the edge of a cube, we consider 2 possible rotations of the end-effector that are 180\degree apart.}
    \label{fig:cube_grasps}
    \vspace{-0.25cm}
\end{figure}

We consider four predefined grasp poses for each face and two grasp poses on each edge of the cube for a total of $48$ possible grasp poses on the cube object. Each grasp pose is defined in the coordinate frame of the cube object. The grasp poses for one face and one edge of the cube is shown in Fig~\ref{fig:cube_grasps}. These poses are repeated for each face/edge. We heuristically choose between the defined grasp poses according to the rules below.

\begin{figure}
    \centering
    \includegraphics[width=0.3\textwidth]{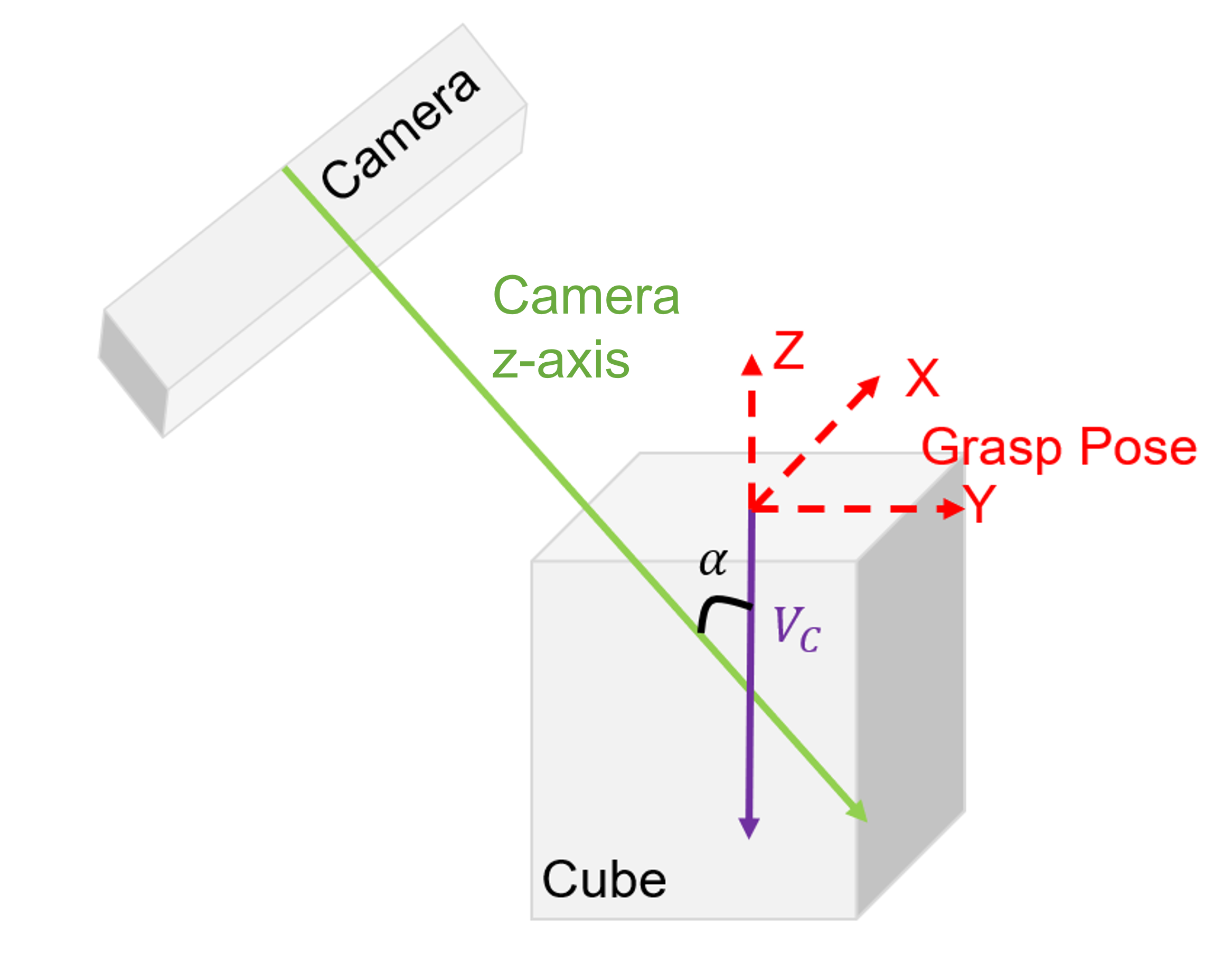}
    \caption{The angle ($\alpha$) between the $z$-axis of the camera and $V_c$ must be less $120\degree$.}
    \label{fig:cubefigure}
    \vspace{-0.25cm}
\end{figure}

\begin{enumerate}
    \item The angle ($\alpha$) between the vector pointing in the direction of the center of the cube from a grasp pose ($V_c$) and the $z$-axis of the end-effector mounted camera must be less than $120\degree$. This is shown in Fig. \ref{fig:cubefigure}.
    \item We choose the face/side of the cube with the largest $z$ coordinate in the world frame.
    \item We filter between the two or four grasp poses (depending on if grasping on a face or side), removing kinematically infeasible grasp poses. If no grasp poses are kinematically feasible, we choose the next highest face/side and repeat. 
    \item From the remaining grasp poses, we choose the pose with the minimum angular difference between the current pose and the respective grasp pose.
\end{enumerate}

The grasp pose is updated in real-time until the user starts the handover. The grasp pose is then fixed, and the robot arm moves towards the selected grasp pose. A bias in the $z$ coordinate is applied to increase the likelihood of the previous grasp (due to Rule 2) and a moving average filter (window size = 5) is applied to the grasp pose. These changes aimed to stabilize the visualization of the grasp pose.



\subsection{Robot Motion Control}

We use resolved-rate motion control to drive the robot towards the desired grasp pose on a linear path. The joint velocity commands are sent to the low-level controller of the Panda robot. Once the robot reaches the grasp pose, the gripper is closed until a force threshold is reached, the end-effector is moved towards a pre-defined dropping location, and the gripper is opened. 

We do not consider the human hand while servoing to the object, however, for safety, we slow the arm down significantly for the last $7$cm of the handover. This ensures there is plenty of time for both the human interaction partner and experiment supervisor to react to a potentially unsafe situation. Additional safety considerations, similar to~\cite{handovers_patrick} could be added, however, this was not the primary focus of this work.

\subsection{Visualization in Augmented Reality}
\label{subsec:vis}

Two 3D models are displayed through AR, to inform the user where the robot "thinks" the object is located and which pose the robot will use to grasp the object. The two models are:

\begin{enumerate}
    \item Detected Object Pose: The wireframe of the object is visualized at its estimated 6D pose. 
    \item Planned Grasp Pose: The 3D model of the robot gripper is visualized as grasping the object wireframe, with the selected grasp pose the robot is planning to execute. The gripper model is visualized with low opacity in order not to block the user's view. 
\end{enumerate}

Once the user initiates the handover with a voice command, the robot's chosen grasp pose and visualizations are fixed in place until the object is taken by the robot from the human.




\section{Hypotheses}
\label{sec:hypothese}

Following our previous work in human-to-robot handovers~\cite{handovers_patrick}, we anticipate that humans will feel safer and more trusting of the robot with the ability to visualize the robot's intent. We also expect the visualization to decrease the number of failed handovers, with less reliance on the experimenter manually stopping the robot. We formulate the following hypotheses to test on a user study with a robot: 

\begin{enumerate}[label=\textbf{H\arabic*}]
    \item The use of \textbf{AR} will have a positive effect on the subjective metrics related to \textbf{fluency} when completing human-to-robot handovers.
    \item The use of \textbf{AR} will have a positive effect on the subjective metrics related to \textbf{trust} when completing human-to-robot handovers.
    \item The use of \textbf{AR} will have a positive effect on the subjective metrics relating to the \textbf{predictability} between the human and the robot.
    \item The use of \textbf{AR} will have a positive effect on the subjective metrics related to \textbf{safety} when completing human-to-robot handovers.
    \item The use of \textbf{AR} will decrease the \textbf{mental load} to complete the task.
    \item \textbf{AR} will have a larger positive effect on all subjective metrics when the simulated error artifact is introduced. 
\end{enumerate}

\section{User Study Design}
\label{sec:userstudy}

A user study was conducted to test the effect of AR on human-to-robot handovers. The methodology for this user study is inspired by our previous work on human-robot collaboration\cite{shray}.

\subsection{Independent Variables}

We manipulate two independent variables:
\begin{enumerate}
    \item Visualization Mode: One of the two following conditions is chosen. 
    \begin{enumerate}
        \item \textbf{AR}: The participant wore the AR headset. The object's 6D pose and the robot's planned grasp pose is visualized, as described in Sec. \ref{subsec:vis}.
        \item \textbf{No AR}: The participant still wore the AR headset, but no visualizations were displayed.
    \end{enumerate}
    \item Presence of Perception Error Artifact: One of the two following conditions is chosen.
        \begin{enumerate}
        \item \textbf{Without ``Simulated Error"}: The object pose estimation result and the selected grasp pose is visualized based on the computed location of the object.
        \item \textbf{With ``Simulated Error"}: We introduce an randomized error to the object pose estimation similar to that experienced in markerless object tracking. The robot servos towards the erroneous grasp pose and the user is expected to compensate for the error by moving the object into the robot gripper. To add error in the object pose, we add error in to the position and orientation separately. The random error added to the object position is parameterized as an angle (sampled uniformly) and a distance which is sampled from a normal distribution ($\mu$=10cm, $\sigma$=1cm) at the start of a handover. The positional error is added along the plane perpendicular to the axis emanating from the camera's forward direction. Random error is also added to the object orientation, by adding noise to each Euler angle of the object, sampled from a normal distribution ($\mu$=10\degree,$\sigma$=3\degree). We expect the handover partner to compensate for these errors by moving the object to the final position of the robot.
    \end{enumerate}    
    The use of simulated error serves two purposes. First, since we use an idealized pose estimator using artificial markers, the simulated error mimics a marker-less pose estimation system by adding a controlled artificial error. Second, we are interested in understanding whether AR would help participants compensate for the robot's errors and whether AR would still bring value in the presence of possible robot perception errors.
\end{enumerate}
We adopt a 2 by 2 design, therefore, each participant experienced 4 different conditions: 
\begin{enumerate} 
    \item \textbf{AR, Without Simulated Error}
    \item \textbf{No AR, Without Simulated Error}
    \item \textbf{AR, With Simulated Error}
    \item \textbf{No AR, With Simulated Error}
\end{enumerate}

To reduce order effects, the order of visualization mode is counterbalanced between participants. The order of \textbf{With Simulated Error} and \textbf{Without Simulated Error} conditions were fixed, however, with the \textbf{With Simulated Error} condition always following the \textbf{Without Simulated Error} condition. This is because we are interested in capturing participants' actual opinions on our proposed system first. Exposing participants to the erroneous robot behavior first would have likely affected their subjective opinion of the overall system.

\subsection{Participant Allocation}

We recruited 16 participants ($13$ male, $3$ female) from within Monash University\footnote[1]{Due to the ongoing COVID-19 pandemic, no external participants could be recruited. This study has been approved by the Monash University Human Research Ethics Committee (Application ID: 27499)}, aged $21-27$ (M = 23.5, SD = 1.82). The participants were not compensated for their time. 13 of the participants had some prior experience with robots whereas 3 had not seen a robot before. 6 of the participants had previous experience with AR, while 7 only heard about AR through popular media. 

\subsection{Procedure}

The experiment took place at a university laboratory under the supervision of an experimenter. Participants stood at a designated location in front of the robot arm to handover objects. Users first read the explanatory statement and signed a consent form. Next, the experimenter explained the experiment by reading from a script. Users then completed a small demographic survey, before completing a training phase without the use of AR. During training, the participant completed a handover without the use of AR, the supervisor manually initiated the handover when the user verbally indicated they were ready. This was repeated as many times as needed for the user to feel comfortable with the robot's behavior.

After training, the user was required to successfully handover the cube 3 times for each condition, for a total of 12 successful handovers during the 4 trials. Finally, the user completed two types of surveys: a survey after each of the 4 test conditions, and one post-experiment survey for additional comments on the study. The survey questions are shown in Table~\ref{tab:questions}. The mean experiment duration was approximately 30 minutes.

\begin{table}[h!]
\centering
  \begin{tabular}{|l|}
    \toprule
     \textbf{Human-Robot Fluency} \\
    Q1: The robot contributed to the fluency of the interaction \\
     \textbf{Trust in Robot}\\
    Q2: I trusted the robot to do the right thing at the right time.\\
    \textbf{Predictability}\\
    Q3: I understand what the robot’s goals are\\
    \textbf{Safety} \\
    Q4: I felt safe completing the handovers\\
    \textbf{Mental Load}\\
    Q5: How mentally demanding was the task? (R) \\
    \bottomrule
    \end{tabular}
    \caption{User Study Survey Questions. (R) indicates a reverse scale.}
    \label{tab:questions}
    \vspace{-0.3cm}
\end{table}

\subsection{Dependent Variables}

\textbf{Subjective Measures:} We adopt the metrics proposed by~\cite{ortenzi2021object} and use a subset of questions that were relevant to the study. Question 5 was added about the cognitive load, adapted from NASA TLX~\cite{Hart1988DevelopmentON}. These questions were designed to measure \textbf{H1} - \textbf{H5}. All questions were measured on a 5 point Likert scale. For Q1-Q4, 1 represents \textit{Strongly disagree} and 5 represents \textit{Strongly agree}. For Q5, 1 represents \textit{Very easy} and 5 represents \textit{Very difficult}.

\textbf{Objective Measures:} We count how many handover failures are experienced by the user. 


\section{Results}
\label{sec:results}

\begin{figure*}[t]
    \centering
    \begin{tikzpicture}
    \definecolor{r}{RGB}{202,0,32}
    \definecolor{rw}{RGB}{213, 123, 111}
    \definecolor{w}{RGB}{198, 198, 198}
    \definecolor{bw}{RGB}{127, 154, 188}
    \definecolor{b}{RGB}{5,113,176}

    \pgfplotsset{
       /pgfplots/bar  cycle  list/.style={/pgfplots/cycle  list={%
            {r!75!black,fill=r!60!white,mark=none},%
            {rw!75!black,fill=rw!60!white,mark=none},%
            {w!75!black,fill=w!60!white,mark=none},%
            {bw!75!black,fill=bw!60!white,mark=none},%
            {b!75!black,fill=b!60!white,mark=none},%
         }
       },
    }
    
    \tikzstyle{every node}=[font=\small]

    
        
        
        
        
        
    

    \begin{axis}[
        name=mainplot,
        xbar stacked,
        title=\textbf{Fluency},
        nodes near coords,
        bar width=0.8,
        width = 0.36\textwidth,
        height = 0.225\textwidth,
        xmin = 0, xmax = 16,
        enlarge y limits={abs=10pt},
        ytick={0,1,2.5,3.5},
        yticklabels={No Err., Err., AR+Err., AR.,},  
        xtick={0,4,8,12,16}, 
        xticklabels={0\%,25\%,50\%,75\%,100\%},         
        legend style={at={(-10,-0.20)}, anchor=north, legend columns=-1, /tikz/every even column/.append style={column sep=0.5cm}},
    ]
    
        \addplot coordinates
        {(3,3.5) (0,2.5) (3,1) (0,0)};
        
        \addplot coordinates
        {(2,3.5) (1,2.5) (5,1) (1,0)};
                
        \addplot coordinates
        {(3,3.5) (4,2.5) (3,1) (7,0)};
                
        \addplot coordinates
        {(5,3.5) (6,2.5) (5,1) (4,0)};

        \addplot coordinates
        {(3,3.5) (5,2.5) (0,1) (4,0)};

    \end{axis}
    \begin{axis}[
        name=secondplot,
        title=\textbf{Trust},
        at={(mainplot.north east)},
        xshift=0.75cm,
        anchor=north west,    
        xbar stacked,
        nodes near coords,
        bar width=0.8,
        width = 0.36\textwidth,
        height = 0.225\textwidth,
        xmin = 0, xmax = 16,
        enlarge y limits={abs=10pt},
        ytick={0,1,2.5,3.5},
        yticklabels={,,,},  
        xtick={0,4,8,12,16},
        xticklabels={0\%,25\%,50\%,75\%,100\%},         
        legend style={at={(0.5,-0.20)}, anchor=north, legend columns=-1, /tikz/every even column/.append style={column sep=0.5cm}},
    ]
    
        \addplot coordinates
        {(0,3.5) (0,2.5) (3,1) (1,0)};
        
        \addplot coordinates
        {(3,3.5) (1,2.5) (3,1) (3,0)};
        
        \addplot coordinates
        {(2,3.5) (2,2.5) (5,1) (4,0)};
        
        \addplot coordinates
        {(6,3.5) (8,2.5) (4,1) (6,0)};
        
        \addplot coordinates
        {(5,3.5) (5,2.5) (1,1) (2,0)};
        
    \end{axis}
    \begin{axis}[
        name=thirdplot,
        title=\textbf{Predictability},
        at={(secondplot.north east)},
        xshift=0.75cm,
        anchor=north west,      
        xbar stacked,
        nodes near coords,
        bar width=0.8,
        width = 0.36\textwidth,
        height = 0.225\textwidth,
        xmin = 0, xmax = 16,
        enlarge y limits={abs=10pt},
        ytick={0,1,2.5,3.5},
        yticklabels={,,,},  
        xtick={0,4,8,12,16},
        xticklabels={0\%,25\%,50\%,75\%,100\%},         
        legend style={at={(0.5,-0.20)}, anchor=north, legend columns=-1, /tikz/every even column/.append style={column sep=0.5cm}},
    ]
    
        \addplot coordinates
        {(1,3.5) (0,2.5) (2,1) (1,0)};
        
        \addplot coordinates
        {(0,3.5) (1,2.5) (2,1) (4,0)};
        
        \addplot coordinates
        {(1,3.5) (0,2.5) (3,1) (1,0)};
        
        \addplot coordinates
        {(6,3.5) (1,2.5) (4,1) (3,0)};
        
        \addplot coordinates
        {(8,3.5) (14,2.5) (5,1) (7,0)};
        
    \end{axis}
    \begin{axis}[
        at={(mainplot.below south west)},
        title=\textbf{Safety},
        yshift=-1cm,
        xshift=3cm,
        anchor=north west,
        xbar stacked,
        nodes near coords,
        bar width=0.8,
        width = 0.36\textwidth,
        height = 0.225\textwidth,
        xmin = 0, xmax = 16,
        enlarge y limits={abs=10pt},
        ytick={0,1,2.5,3.5},
        yticklabels={No Err., Err., AR+Err., AR.,},  
        xtick={0,4,8,12,16},
        xticklabels={0\%,25\%,50\%,75\%,100\%},         
        legend style={at={(0.5,-0.20)}, anchor=north, legend columns=-1, /tikz/every even column/.append style={column sep=0.5cm}},
    ]
        \addplot coordinates
        {(0,3.5) (0,2.5) (1,1) (0,0)};
        
        \addplot coordinates
        {(1,3.5) (0,2.5) (1,1) (2,0)};
        
        \addplot coordinates
        {(1,3.5) (3,2.5) (4,1) (2,0)};
        
        \addplot coordinates
        {(5,3.5) (6,2.5) (6,1) (8,0)};
        
        \addplot coordinates
        {(9,3.5) (7,2.5) (4,1) (4,0)};

    \end{axis}
    \begin{axis}[
        title=\textbf{Mental Load (R)},
        at={(secondplot.below south west)},
        yshift=-1cm,
        xshift=3cm,
        anchor=north west,    
        xbar stacked,
        nodes near coords,
        bar width=0.8,
        width = 0.36\textwidth,
        height = 0.225\textwidth,
        xmin = 0, xmax = 16,
        enlarge y limits={abs=10pt},
        ytick={0,1,2.5,3.5},
        yticklabels={,,,},   
        xtick={0,4,8,12,16},
        xticklabels={0\%,25\%,50\%,75\%,100\%},         
        legend style={at={(0.5,-0.35)}, xshift=-3cm, anchor=north, legend columns=-1, /tikz/every even column/.append style={column sep=0.5cm}},
    ]
    
        \addplot coordinates
        {(5,3.5) (8,2.5) (2,1) (4,0)};	
        
        \addplot coordinates
        {(9,3.5) (7,2.5) (8,1) (12,0)};	
        
        \addplot coordinates
        {(1,3.5) (1,2.5) (5,1) (0,0)};	
        
        \addplot coordinates
        {(1,3.5) (0,2.5) (1,1) (0,0)};	
        
        \addplot coordinates
        {(0,3.5) (0,2.5) (0,1) (0,0)};

        \legend{1 (Strongly Disagree), 2, 3, 4, 5 (Strongly Agree)}
        
    \end{axis}
    
        
        
        
        
        

\end{tikzpicture}
    \\
    \caption{Summary of the user study results. Answers are on 5-Likert scale and Mental Load is on a reverse scale, indicated by (R)}
\label{fig:results}
\end{figure*}
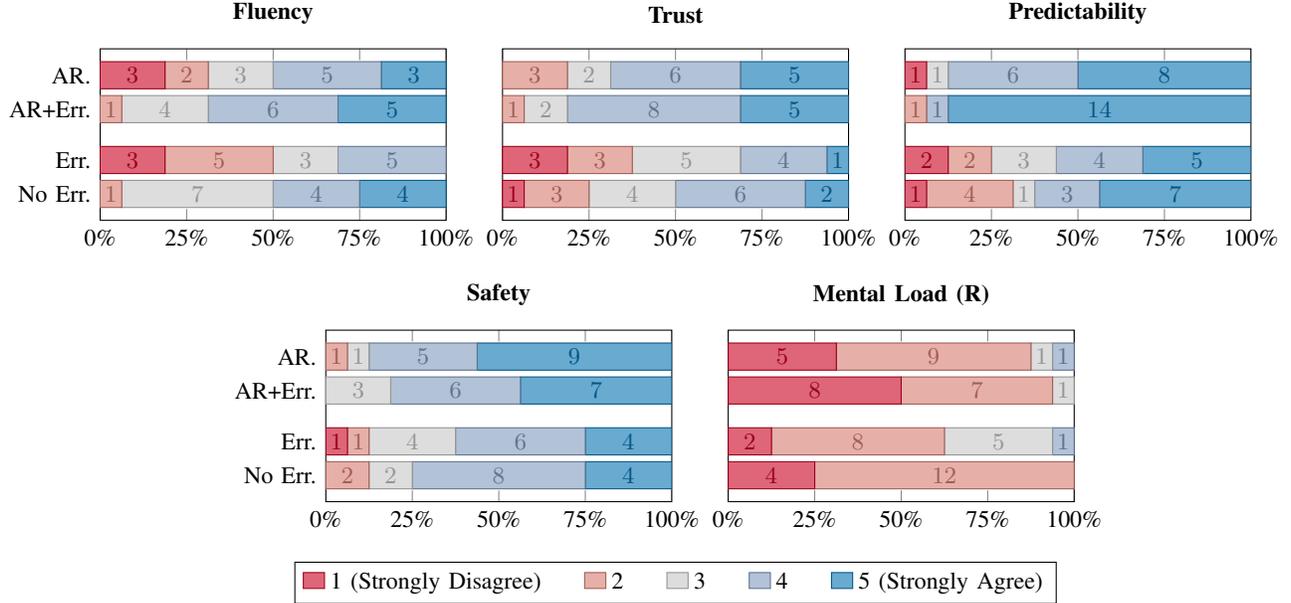

We study the effect of the independent variables on the dependent variables. In total we analyze $(N = 16) * 4 = 64$ conditions. The results of the user study are shown in Fig.~\ref{fig:results}.

\subsection{Objective Measures: Handover Failures}



The only failure mode observed during this study was the robot being unsuccessful in grasping the object. This was primarily due to the person not compensating for the slight noise in the object pose estimation for the \textbf{Without Simulated Error} condition or the additional error in the \textbf{With Simulated Error} condition. The number of handover failures are shown in Tab.~\ref{tab:failures}. As expected, the most number of failures occurred during the \textbf{With Simulated Error} and \textbf{No AR} case. This was due to users being unable to correctly estimate the grasp pose quickly enough. Overall, we observe that \textbf{AR} led to fewer number of handover failures.

\begin{table}[]
\centering
\begin{tabular}{l|ll}
        Number of Failures         & \textbf{No AR} & \textbf{AR} \\ \hline
Without Simulated Error   & 4     & 4  \\
    With Simulated Error & 10    & 4 
\end{tabular}
\caption{The distribution of failed handovers in all experiments. Each condition had a total of 48 successful handovers.}  
\label{tab:failures}
\end{table}


\subsection{Subjective Measures: Survey Questions}


For each question, we perform an Aligned Rank Transform~\cite{ART}, which allows for nonparametric testing of interactions and main effects of ordinal data, such as a Likert Scale, using standard ANOVA techniques. Interaction effects occur when the effects of an independent variable depend on the other variable. In this study, none of the questions had a significant interaction between the independent variables, which allowed us to examine the significance of an independent variable and collapsing the other independent variable. The comparison between \textbf{AR} and \textbf{No AR} cases is shown in Tab.~\ref{tab:arvsnoa}, and the comparison between \textbf{With Simulated Error} and \textbf{Without Simulated Error} is shown in Tab.~\ref{tab;errorvsnoerr}.



\setlength{\tabcolsep}{1.96mm}
\begin{table}[h!]
\begin{tabular}{|l|l|l|l|l|l|l|}
\hline
                    & \multicolumn{2}{l|}{AR} & \multicolumn{2}{l|}{No AR} & \multirow{2}{*}{F(1,63)} & \multirow{2}{*}{p} \\ \cline{1-5}
                    & $\mu$     & $\sigma$    & $\mu$      & $\sigma$      &                          &                    \\ \hline
Q1 (Fluency)        & 3.56         & 1.24     & 3.16          & 1.27       & 2.710                    & 0.105              \\ \hline
Q2 (Trust)          & 3.94         & 0.98     & 3.06          & 1.19       & 11.078                   & \textbf{0.001}     \\ \hline
Q3 (Predictability) & 4.50         & 0.95     & 3.59          & 1.41       & 6.083                    & \textbf{0.017}     \\ \hline
Q4 (Safety)         & 4.31         & 0.81     & 3.78          & 1.04       & 5.157                    & \textbf{0.027}     \\ \hline
Q5 (Mental Load)    & 1.72         & 0.73     & 2.03          & 0.69       & 5.488                    & \textbf{0.022}     \\ \hline
\end{tabular}
\caption{Results from the user study comparing AR and No AR and collapsing the other variable. The significance was calculated using Aligned Rank Transform ANOVA. Q2 - Q5 (bold) all have significant results.}
\label{tab:arvsnoa}
\end{table}

\setlength{\tabcolsep}{1.8mm}
\begin{table}[h!]
\begin{tabular}{|l|l|l|l|l|l|l|}
\hline
                    & \multicolumn{2}{l|}{Without Error} & \multicolumn{2}{l|}{With Error} & \multirow{2}{*}{F(1,63)} & \multirow{2}{*}{p} \\ \cline{1-5}
                    & $\mu$        & $\sigma$       & $\mu$      & $\sigma$      &                          &                    \\ \hline
Q1 (Fluency)        & 3.81            & 0.93        & 2.91          & 1.30       & 8.530                    & \textbf{0.005}     \\ \hline
Q2 (Trust)          & 3.69            & 1.06        & 3.31          & 1.26       & 1.955                    & 0.167              \\ \hline
Q3 (Predictability) & 4.22            & 1.26        & 3.88          & 1.29       & 1.143                    & 0.289              \\ \hline
Q4 (Safety)         & 4.06            & 0.88        & 4.03          & 1.06       & 0.001                    & 0.979              \\ \hline
Q5 (Mental Load)    & 1.66            & 0.55        & 2.09          & 0.82       & 5.947                    & \textbf{0.018}             \\ \hline
\end{tabular}
\caption{Results from the user study comparing With Simulated Error and Without Simulated Error and collapsing the other variable. The significance was calculated using Aligned Rank Transform ANOVA. Q1 and Q5 have significant results.}
\label{tab;errorvsnoerr}
\end{table}

\textbf{Fluency (Q1, H1)}:  \textbf{AR} has a positive effect on the subjective measure of fluency, increasing the mean rating from $3.16$ to $3.56$, however the difference was not statistically significant. The presence of \textbf{Simulated Error}, however, significantly reduced the fluency of the interaction. Therefore, \textbf{H1} was not affirmed.

\textbf{Trust (Q2, H2)}: The mean subjective perception of trust increases from $3.06$ to $3.94$ with the use of \textbf{AR} and the difference was statistically significant. This affirms \textbf{H2}. We do not observe a significant difference between the \textbf{With Simulated Error }and \textbf{Without Simulated Error} conditions. 

\textbf{Predictability (Q3, H3)}: The subjective level of predictability significantly increases from $3.59$ to $4.50$ with the use of \textbf{AR}. This affirms \textbf{H3}. Moreover, most participants noted that during the \textbf{Without Simulated Error} and \textbf{AR} condition that they fully understood the robot's goals.

\textbf{Safety (Q4, H4)}: Use of AR significantly increases the subjective level of perceived safety, increasing the mean rating from $3.78$ to $4.31$. This affirms \textbf{H4}. 


\textbf{Mental Load (Q5, H5)}: The subjective mental load significantly decreases through the use of \textbf{AR}, from $2.03$ to $1.72$. This affirms \textbf{H5}. Fig \ref{fig:results} (Mental Load) shows that the mental load is similar in all conditions apart from \textbf{With Simulated Error} and \textbf{No AR}, where the user experienced an increase in the mental load. Furthermore, we found that the introduction of \textbf{Simulated Error} caused the mental loads of the participants to increase significantly. 




\textbf{Value of AR under Pose Estimation Errors (H6):} We are interested in investigating if the the benefit of using AR is more pronounced when there is~\textbf{Simulated Error} imposed on the object pose. For each question we examine the difference between \textbf{AR} and \textbf{No AR} for both \textbf{With Simulated Error} and \textbf{Without Simulated Error} conditions. This formed two additional variables for each question. We then used the Kruskal-Wallis nonparametric test, which can be used for skewed Likert data~\cite{SchrumHRI}. This test aims to see if \textbf{AR} has a larger effect on the performance of the system in the \textbf{With Simulated Error} condition compared to the \textbf{Without Simulated Error} condition. The results of the Kruskal-Wallis test are summarized in Tab.~\ref{tab:kw}. Q1 (Fluency) and Q4 (Safety) have significant results, which partially affirms \textbf{H6}. Therefore, we observe that the advantage of using AR is more pronounced in metrics related to perceived fluency and safety when the robot's object pose estimation is imperfect.

\begin{table}[]
\centering
\begin{tabular}{l|ll}
   & H(1)  & P              \\ \hline
Q1 (Fluency) & 6.960 & \textbf{0.003} \\
Q2 (Trust) & 1.114 & 0.324          \\
Q3 (Predictability) & 0.051 & 0.808          \\
Q4 (Safety) & 3.767 & \textbf{0.037} \\
Q5 (Mental Load) & 1.114 & 0.244         
\end{tabular}
\caption{We check if AR had a greater improvement on results in the presence of the Simulated Error Artifact. A Kruskal-Wallis test was performed for each question. Significant results are indicated in bold text.}
\label{tab:kw}
\end{table}

\section{Discussion}
\label{sec:discussion}

The user study results suggest that hypotheses \textbf{H2}, \textbf{H3}, \textbf{H4}, \textbf{H5} were affirmed. No significant results were observed for \textbf{H1}. \textbf{H6} was partially affirmed only for the subjective metrics of perceived fluency and safety.


The users' perceived level of safety was similar between both \textbf{Without Simulated Error} and \textbf{With Simulated Error} conditions, as shown in Fig.~\ref{fig:results} (Safety), with a rating of $4.03$ and $4.06$ respectively. This suggests that overall, users felt safe interacting with the system.

Participants discussed the usefulness of seeing the robot's goal as it allowed them to adjust the object transfer point to account for errors in the robot's goal. Participant 3 noted that \textit{``Visualizing the gripper was probably the best thing, since I was able to adjust my position to have the robot grasp the object and handover."}. Further, Participant 4 said that \textit{``it was far more confusing when I did not have any grasp information"}. This is further shown by a large difference ($0.91$) in the subjective rating of the predictability between the \textbf{No AR} and \textbf{AR} conditions. Participants also noted that this visualization increased their feeling of trust and fluency in their handover. 

For the \textbf{AR} and \textbf{With Simulated Error} condition, it was observed that users tend to adjust their hand position relatively quickly to the final grasp point of the robot. Once this was completed, the users tended to stay quite still until the robot had completed the handover. In contrast, during the \textbf{No AR} and \textbf{With Simulated Error} condition, the users, would continuously adjust their position as the robot moved. This takes more effort from the user and is the most likely cause for the increased mental load to complete the task without \textbf{AR}. Further studies could track the position of the user's wrist to substantiate this explanation.

Some users noted that the visualization could be distracting, especially if it had a slight misalignment with the object for the \textbf{Without Simulated Error} condition. Participant 9 said that \textit{``The offset in the wireframe and the mismatching of visualization to the gripper was distracting at times. But once I got used to it, the visualization was helpful for my understanding of what was happening."}. 

\section{Conclusion and Future Work}
\label{sec:conclusion}

Conveying the robot's intent is often disregarded in human-to-robot handovers. We propose a novel AR-based interface to communicate the robot's internal state to the user by visualizing the estimated pose of the object and where the robot is planning to grasp the object. User studies demonstrate that conducting the proposed interaction through AR significantly improves the subjective experience of the participants in terms of fluency of interaction, trust towards the robot, perceived safety, mental load, and predictability. Our proof-of-concept is achieved for a single object, in which its pose is tracked using artificial markers. The proposed method is subjectively perceived as safe, fluent and trustworthy even when random artificial noise is introduced to the grasp pose. This suggests that the approach would be well-suited to more realistic scenarios where accurate detection of the object may not be possible and that humans are willing to compensate for errors in robotic vision if robots can communicate their intent to their human partners.

This paper serves as a foray into communication methods for object handovers. What information should be visualized, and how remains an open research question. Different visualizations can be explored, such as the robot's future trajectory or an approximate bounding box of the object instead of the full 6D pose. Future directions include the possible use of interactive markers, and using hand gestures to allow the user to possibly move the robot's grasp pose to compensate for errors or grasp objects in a semantically preferred way.




\bibliographystyle{IEEEtranN}
{\footnotesize \bibliography{refs}}

\end{document}